\documentclass[twocolumn]{article}
\def\BibTeX{{\rm B\kern-.05em{\sc i\kern-.025em b}\kern-.08emT\kern-.1667em\lower.7ex\hbox{E}\kern-.125emX}}

\usepackage{amsfonts}
\usepackage{booktabs}
\usepackage{multirow}
\usepackage{tabulary}
\usepackage{tabularx}
\usepackage{graphicx}
\usepackage{subcaption}
\usepackage{amsmath}
\usepackage[colorinlistoftodos]{todonotes}
\usepackage{hyperref}

\begin{document}
\title{Reinforcement Learning for Personalized Dialogue Management}

\author{Floris den Hengst}

\author{Mark Hoogendoorn}

\author{Frank van Harmelen}

\author{Floris den Hengst$^{1}$, Mark Hoogendoorn$^{2}$, Frank van Harmelen$^{2}$, Joost Bosman$^{1}$  \\
        \small $^{1}$ING Bank N.V. \\
        \small $^{2}$Vrije Universiteit Amsterdam \\
}

%

\date{July 25, 2019}
\maketitle

\begin{abstract}
Language systems have been of great interest to the research community and have recently reached the mass market through various assistant platforms on the web. Reinforcement Learning methods that optimize dialogue policies have seen successes in past years and have recently been extended into methods that \emph{personalize} the dialogue, e.g. take the personal context of users into account. These works, however, are limited to personalization to a single user with whom they require multiple interactions and do not generalize the usage of context across users. This work introduces a problem where a generalized usage of context is relevant and proposes two Reinforcement Learning (RL)-based approaches to this problem.
The first approach uses a single learner and extends the traditional POMDP formulation of dialogue state with features that describe the user context. The second approach segments users by context and then employs a learner per context.
We compare these approaches in a benchmark of existing non-RL and RL-based methods in three established and one novel application domain of financial product recommendation. We compare the influence of context and training experiences on performance and find that learning approaches generally outperform a handcrafted gold standard.
\end{abstract}



\providecommand{\keywords}[1]
{
  \small	
  \textbf{\textit{Keywords---}} #1
}


\section{Introduction}
\label{sec:intro}
The use of language by machines has been one of the central challenges in Artificial Intelligence since its initiation as a field of research \cite{turing1950computingmachinery} \cite{mccarthy2006proposal}. Decades of research have advanced the state of art to such an extent that major consumer-facing web platforms currently offer text- and voice-based  `assistant' capabilities, such as Tencent's WeChat, Microsoft's Cortana, Google's Assistant etc. These platforms have made access to the web through dialogue ordinary. Although such platforms offer high-quality Automatic Speech Recognition (ASR), Natural Language Understanding (NLU) and audio synthesis modules, Dialogue Management (DM) modules are typically handcrafted and require many non-trivial decisions in design and implementation. \emph{Learned} DM based on the formalism of Partially Observable Markov Decision Processes (POMDPs) has shown promising results in task-oriented dialogue systems, both in simulation and real-life settings \cite{roy2000spoken} \cite{gavsic2010gaussian} \cite{young2013pomdp}.

Personal context is understood to be fundamental to efficient human-human communication \cite{bergmann2007language}. As a consequence, recent works have addressed the usage of personal context in DM.
For example, \cite{thompson2004personalized}, \cite{mahmood2014dynamic} and \cite{AAAI1816104} used previous interactions with a user to directly estimate that users' preferences and then used these estimates in policy optimization. An alternative approach based on transfer learning was presented in \cite{casanueva2015knowledge}. It requires a similarity metric and weighting regime and performance degrades when these are not available.
None of these methods generalize the usage of context across users and none of them leverages information available prior to some users' first interaction with the system.

We propose two approaches that optimize the DM policies using personal context. Both approaches are based on the POMDP formalism of learned DM. The first approach consists of extending the POMDP state space with features that describe the personal context of the user. The DM module automatically learns how to use this information for both groups, i.e. it learns the task at hand and segmentation of users simultaneously. This approach allows for personalization to emerge gracefully, e.g. only when enough data is present and when the user model is sufficiently informative for personalization. We compare this approach with a method that explicitly segments users and then uses a learner per user segment. The segmentation of interactions with different user groups mitigates the issue of a `mixed' signal but leaves less experiences to learn from per learner.

To test our approaches, we extend an existing benchmark for POMDP-based statistical DM for recommendation in three ways \cite{casanueva2017benchmarking}. Firstly, we add a novel recommendation task in the financial domain. Here, different user groups have different familiarity with products and specify their preferences at different levels of detail as a result. Secondly, we change the user simulator in the benchmark to reflect this scenario. Thirdly, we add three non-POMDP based approaches to the benchmark: a randomized approach, an approach with a task-specific heuristic and a state-of-art approach based on entropy minimization \cite{wu2015entropy}. To the best of our knowledge, this comparison between POMDP and non-POMDP based approaches on task-oriented dialog management is novel.

We use the extended benchmark to investigate when each approach is suitable for personalized DM and we investigate the impact of available data to the achieved level of personalization. We first introduce and formalize the recommendation task in Section~\ref{sec:task} and survey related work in Section~\ref{sec:related_work}. Next, we introduce the generic approach to RL for DM and then introduce our extensions. The experimental setup consists of recommendation in existing and novel domains, a user simulator for personalized DM and a benchmark of POMDP and non-POMDP algorithms, is introduced in Section~\ref{sec:setup}. After describing and analyzing the results in Section~\ref{sec:results}, we conclude with a discussion in Section~\ref{sec:discussion}.

\section{Task Description}
\label{sec:task}
This work addresses DM in task-oriented dialogue systems. These systems aim to solve a task by interacting with the user in a conversational style. A popular task for these systems is to recommend a suitable item for a user. The system elicits user preferences or constraints during a dialogue and recommends items from a given item database. We introduce this task formally.

The task addressed in this paper can be formalized as a $q$-ary two-player interactive search game \cite{pelc2002searching}. In these game, the goal of one player, dubbed Questioner, is to find a target subset $X_{\mathit{target}} \subseteq X = \{x_{1}, \dots, x_n\}$  out of a universe of items $X$ of size $n$ by asking questions to the other player, the Responder. In this case, each $x_i \in X$ consists of a vector of values $\langle x_{i1},  \dots, x_{im} \rangle$ for features $\{f_1, \dots, f_m\}$. $X_{\mathit{target}}$ is identified by a set of constraints $C$, in the form of the desired value $c_j$ for some feature $f_j$. We assume $\forall c_j \in C, \forall x_i \in X_{\mathit{target}}, \forall x_{ij} \in x_i : x_{ij}=c_j$. Each $c_j$ eliminates a part of the search space. We use $C_t$ to denote the set of constraints at game turn $t$ and $X_{C_t}$ to denote the corresponding candidate item set.


Both the typical $q$-ary search game and our variation are generalizations of the R\'enyi-Ulam game (RU game), also known as the binary search game or the parlour game `20 questions'. In RU games, Questions are limited to confirmation of a single constraint, i.e. they are all of the form `$c_j \in C$?' In this format, the optimal question halves the candidate item set $X_{C_t}$ in the optimal case. In our setting, however, the optimal decrease in candidate item set size depends on the distribution of values for all $f_j$'s in $X_{C_t}$. The Questioner may use knowledge about these distributions in selecting a $f_j$ to ask a constraint for. We therefore include a policy that uses knowledge about the distribution of values in all $f_j$'s as a search heuristic. More so, the Responders' tendency to provide constraints for a feature $f_j$ may not be distributed uniformly in realistic settings. A Questioner with access to past plays may use this experience to estimate the likelihood of a constraint for a feature being present to find an item more efficiently. We therefore include approaches that can leverage experience into our benchmark, see Section~\ref{sec:benchmark} for details.

\section{Related Work}
\label{sec:related_work}
Most approaches to personalizing dialogue systems can be categorized as learning-based or rule-based. We provide a brief overview of approaches in both categories. An example of a rule-based approach can be found in \cite{goker2000personalized} and \cite{thompson2004personalized}. This system uses a model of user preferences for constraints $c_j$ to weigh factors that determine similarity of a user query to the items in $X$. The DM policy is handcrafted, which typically entails many nontrivial decisions that can seriously impact system performance \cite{litman2000automatic}. More recent examples, such as \cite{kim2014acquisition}, \cite{bang2015example} and \cite{shum2018eliza} collect user-related facts in a knowledge graph. These facts are then used to personalize hand-crafted response templates. These approaches focus on personalized natural language generation and have handcrafted DM modules.

Learning-based approaches, on the other hand, optimize the DM policy using experiences with real or simulated users. A conversational shopping recommender is described in \cite{mahmood2014dynamic}. It requires multiple interactions with a specific user and has a query-response interaction style. An example with a natural language interaction style based on transfer learning can be found in \cite{casanueva2015knowledge}. It initializes a policy for the target user by training on data from interactions with similar users. The authors find that it is beneficial to include data from dissimilar users, albeit with lower weights, as this results in better coverage of the state space during training. A drawback of the approach is that it requires a suitable similarity metric. A transfer learning-based approach that does not suffer from this drawback is introduced in \cite{AAAI1816104}. A policy is optimized using a global optimization criterion and all available experiences. Next, the optimization criterion is extended with user-specific slot-value preference estimates which are updated in subsequent interactions. This approach only adapts to individual users in terms of slot-value preferences and requires multiple interactions with a single user. A third transfer learning-based approach is presented in \cite{genevay2016transfer}. The selection of experiences to train the model on for a specific user is cast as a multi-armed bandit problem. Finding a source of experiences out of all $n$ users, however, requires at least $n$ bandit trials. This limits applicability to scenarios with a small number of users.

None of the approaches discussed so far leverage information external to the conversation, e.g. context, to optimize the dialogue policy. In non-conversational recommendation, however, numerous works rely on the users' personal contexts. As a full survey is out of scope for this paper, so we focus on generic trends instead. Recommender systems are typically classified as content-based, collaborative filtering or a hybrid of these two. Content-based recommender systems `exploit the user profile to suggest relevant items by matching the profile representation against that of items to be recommended' and thus rely on the users' personal context \cite{pazzani2007content}. Collaborative filtering selects items for recommendation by looking at past consumption patterns by similar users and personal context can be used to determine similarity of users \cite{karatzoglou2010multiverse} \cite{madani2005contextual} \cite{adomavicius2011context}. Out of these approaches, contextual bandit methods are specifically related to this work. These methods aim to determine how elements of personal context affect relevance of items through subsequent interactions with users \cite{li2010contextual}. These methods, however, are not suitable for conversational settings as they do not take sparsity of rewards and the sequential nature of these settings into account.


\section{Approach}
This section describes two novel approaches to personalized DM for the interactive recommendation task described in Section~\ref{sec:task}. First, the formalism of Partially Observable Markov Decision Problems is described and it is explained how it it can be applied to DM for the interactive recommendation task.

\begin{figure}
    \begin{subfigure}[]{\columnwidth}
        \includegraphics[width=\columnwidth]{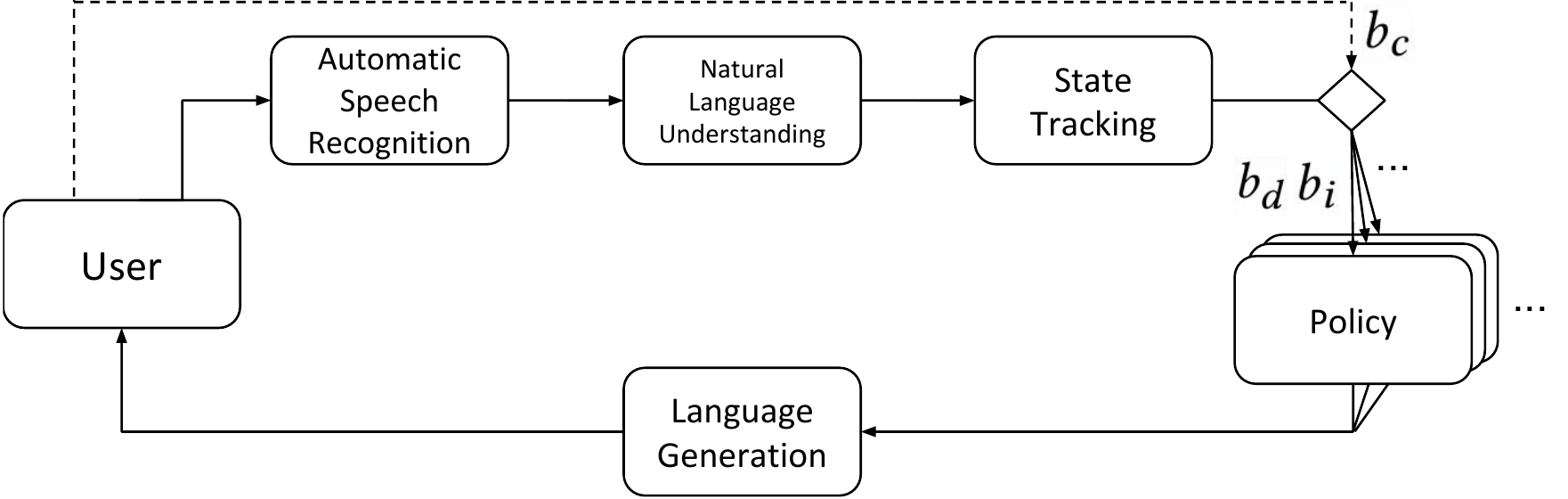}
        \caption{Segmentation-based.}
      \label{fig:split-based}
    \end{subfigure}%
    
    \begin{subfigure}[]{\columnwidth}
        \centering
        \includegraphics[width=\columnwidth]{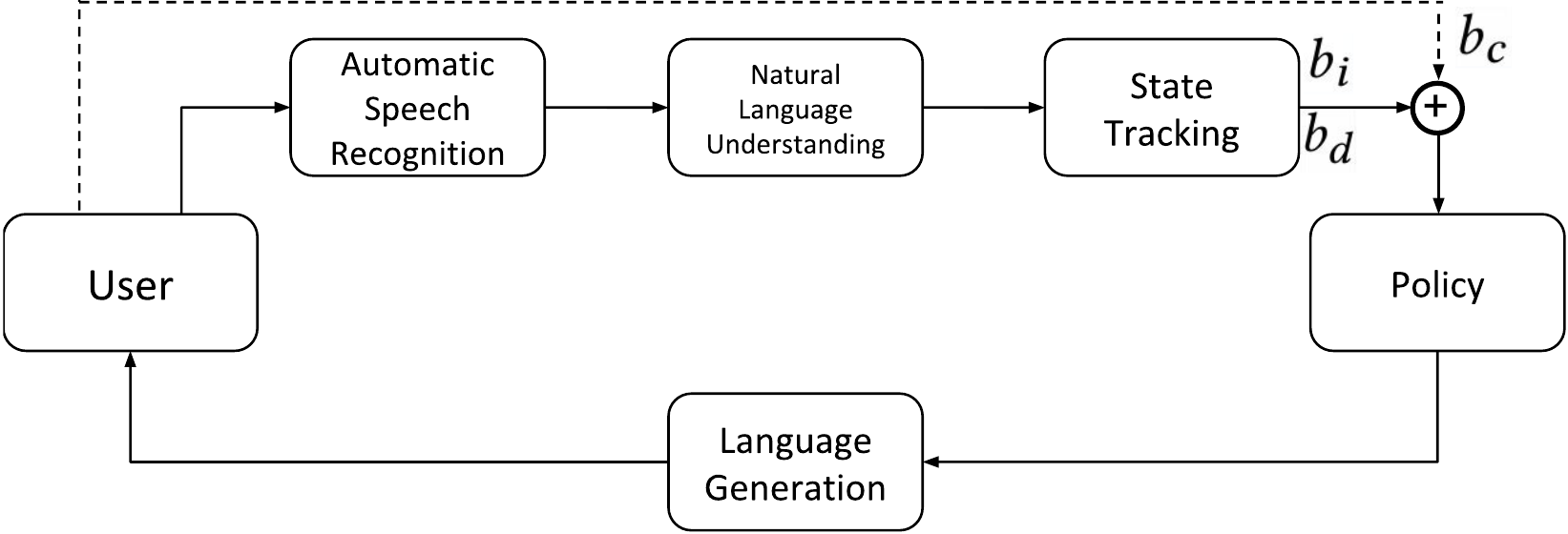}
        \caption{State-based.}
        \label{fig:state-based}
    \end{subfigure}%
    \caption{RL-based approaches to personalized DM.}
    \label{fig:approaches}
\end{figure}

\subsection{RL for DM}
\label{sec:rl4dialogue}
State of the art statistical dialogue systems cast DM as a Partially Observable Markov Decision Problem (POMDP) \cite{roy2000spoken} \cite{williams2007partially}. A POMDP is a generalization of a Markov Decision Process where the true state is not directly observable, but must be estimated through observations. In dialogue systems, the source of uncertainty about the true state stems from errors in Automatic Speech Recognition (ASR) and Natural Language Understanding (NLU) modules. The POMDP is defined as $M = \langle S, A, T, R, \Omega, O \rangle$ where $S \in \{s_1, \dots, s_n\}$ denotes a finite set of partially observable states representing user intentions and dialogue history, $A \in \{a_1, \dots, a_m\}$ is a finite set of actions representing system responses, $T \colon S \times A \times S \to [0, 1]$ is a probabilistic transition function over states and $R \colon S \times A \to \mathbb{R}$ denotes a reward function based on number of turns and accuracy of recommendation, $\Omega \in \{o_1, \dots, o_l\}$ is a finite set of observations available to the system, and $O \colon \Omega \times A \times S \to [0,1]$ denotes a probabilistic function over observations, actions and states. The true state $s$ is unavailable to the agent, only observations $\Omega$ are.

The dependence of $O$ on $\Omega$ and $A$ makes the decision process non-Markovian and thus unsuitable for standard RL algorithms. The Markovian property can be regained, however, by maintaining a Bayesian belief over $S$ and substituting the original state space with this belief space. This substitution leaves us with a continuous MDP with an input space $B \in \{b_1, \ldots, b_{o}\}$ with dimensionality $|S| - 1$, which is too complex for most practical purposes. In practice, however, the belief space can be significantly reduced in size by splitting it into factors and assuming mutual independence between factors. In dialogue systems aimed at the interactive recommendation task from Section~\ref{sec:task}, the belief space can be split into a factored belief space $B'$ consisting of dialogue history belief $b_d$ and a user intention belief $b_i$. The dialogue history $b_d$ describes, for example, whether the system has already recommended an item $x_i$ or requested a constraint for feature $f_j$. The user intention belief describes preferences of the user w.r.t. the product database. Maintaining this state is a challenge in itself, but outside of the scope of this work. See \cite{williams2013dialog} and \cite{henderson2014second} for overviews. As $B$ is replaced by $B'$ and not used anymore, we denote $B'$ as $B$ from here on. 

Constructing the POMDP involves some design decisions based on the task at hand. Specifically, $A$ should contain actions that are useful or necessary for the agent to achieve its task. For the interactive recommendation task the agent plays the part of Questioner. The available utterances should thus at least reflect requesting a constraint for each feature $f_j$ and recommending an item. Additional actions can make the dialogue more natural and efficient, such as confirmation questions of the form `$c_j \in C?$' and selection questions of the form `$c_j \in C~or~c_{j'} \in C?$'.

Besides a suitably defined $A$, the POMDP should be constructed with an $R$ that reflects the goal of the task at hand. This work is based on a benchmark further described in Section~\ref{sec:setup}. In the benchmark $R$ is defined as
\begin{equation}
    \label{eq:reward}
    20 * acc(X_{target}, \langle a^1, \dots, a^l\rangle) - l
\end{equation} for a given $X_{target}$ and trajectory of system actions $\langle a^1, \ldots, a^l \rangle$ of length $l$. $acc()$ returns $1$ if the trajectory contains a recommendation action for an item $x_i \in X_{target}$ and $0$ otherwise. The goal is to find the optimal function $\pi^*: B \to A$ that maximizes the expected sum of discounted future rewards
\begin{equation}
    \label{eq:policy}
    \pi^{*}(b) = \underset{a}{\mathrm{arg~max}} Q^{\pi^*}(b, a), \forall b \in B, \forall a \in A    
\end{equation}
where
\begin{equation}
    \label{eq:q_function}
    Q^{\pi^{*}}(b, a) = E_{\pi^*} \Big\{\ \sum_{k=0}^{\infty}\gamma^k r^{t+k+1}|b^{t} = b, a^{t} = a \Big\}
\end{equation}
and $\gamma \in [0,1]$ is a factor weighing future rewards and $b^t$ and $a^t$ are future beliefs and actions.

\subsection{Personalized Dialogue Management}
We present two approaches to DM using personal context of the user based on the formalism described. Figure~\ref{fig:approaches} provides an overview of the two methods. Both use a vector describing the agents' belief of \emph{personal context} $b_c$ of the user to optimize the dialogue for specific users. This may include any available information about the user that may aid in policy optimization. Examples of context include demographics, purchase history and previous interactions. Note that context need not be constant during or in between dialogues. This section describes how context is used in both methods.

The method in Figure~\ref{fig:split-based} is based on segmentation of the user population by context. It assumes a function $M : B_c \to G$ that maps agent beliefs on user contexts $B_c \in \{b_{c_1}, b_{c_2}, \ldots, b_{c_n}\}$ to segments $g \in G$ ($g$ for `group'). A separate policy $\pi_{g}(b_d, b_i)$ is maintained that exclusively interacts with contexts $b_c$ for which $M(b_c) = g$. As the policy interacts with user contexts in a single segment, it learns a policy optimal for that segment using only beliefs on dialogue history $b_d$ and user intentions $b_i$. The context $b_c$ is not available to the policy. A benefit of this approach is the absence of negative transfer between segments: behaviors suitable to only a particular segment of users are only learned by that segments' policy and will not be considered suitable by policies serving the other segments. On the other hand, there cannot be any positive transfer either: each policy is exposed to less interactions which may result in poor belief state space coverage and degraded performance. Furthermore, it may be nontrivial to find a suitable segmentation function $M$ as this involves finding an unambiguous context representation and determining the number of segments.

The method in Figure~\ref{fig:state-based} does not suffer from these drawbacks. It consists of concatenating beliefs on dialogue history $b_d$, user intentions $b_i$ and context $b_c$. The resulting belief vector is then used as input to a single policy $\pi_p(b_d, b_i, b_c)$ for the entire user population. An algorithm that optimizes $\pi_p$ now jointly learns DM and the usage of context therein. This allows for the learner to only use context when it is beneficial and liberates us from defining segmentation or similarity criteria. The composed learning task, however, may be significantly more challenging as users from different segments may have conflicting desires. This might lead to a form of negative transfer that the algorithm optimizing $\pi_p$ has to be robust to which may require more training data.

\begingroup
\setlength{\tabcolsep}{2pt} 
\begin{table}[t!bph]
    \centering
    \begin{tabulary}{\columnwidth}{ c c C C }
         Domain
         &
         \# Items
         &
         Group 1 \& 2
         &
         Group 2 only
         \\ \hline
         CR
         &
         110
         &
         price range
         &
         area, food
         
         \\ \hline
         \multirow{3}{*}{SFR}
         &
         \multirow{3}{*}{271}
         &
         price range, allowed for kids, good for meal
         &
         area, near, food
         \\ \hline
         
         \multirow{5}{*}{LAP}
         &
         \multirow{5}{*}{123}
         &
         utility, price range, weight range, warranty, is for business computing
         &
         family, processor class, sys memory, platform, drive range, battery rating
         \\ \hline
         
         \multirow{5}{*}{FIN}
         &
         \multirow{5}{*}{14}
         &
         minimum age, purpose, account
         &
         name, insurance, max. duration, min. duration, max. principal, min. principal
         \\ \hline
    \end{tabulary}
    \caption{Usage of slots for constraints for the two user groups. Group 1 denotes users unfamiliar with the domain or `laypersons' while Group 2 denotes users experienced in the domain or `experts'. Expert users always use three constraints, whereas layperson users have between one and three constraints.}
    \label{tab:error_slots}
\end{table}

\section{Experimental Setup}
\label{sec:setup}
The goal of this paper is to evaluate the proposed approaches for personalized dialogue management. We split this goal into the following research questions. In a personalized DM task,
\begin{enumerate}
    \item[Q1] when do learning-based algorithms outperform handcrafted algorithms?
    \item[Q2] when do belief state-based approaches outperform segmentation-based approaches?
    \item[Q3] how well do existing approaches generalize to the novel domain of financial product recommendation?
\end{enumerate}
Regarding these research questions, we hypothesize:
\begin{enumerate}
    \item[H1] learning-based approach only outperform handcrafted approaches in the presence of preprocessing errors.
    \item[H2] belief state-based approaches perform comparable to or better than segmentation-based approaches.
    \item[H3a] in the new domain, learning-based approaches perform comparable to existing domains.
    \item[H3b] in the new domain, handcrafted approaches perform worse than in existing domains.
\end{enumerate}

The experimental setup is based on a benchmark suite for task-oriented dialog management \cite{casanueva2017benchmarking}. The suite includes a user simulator, a dialog management module and DM algorithms. The benchmark further consists of recommendation tasks in three domains: recommendation of restaurants in Cambridge (CR), of restaurants in San Francisco (SFR) and laptops (LAP), we refer to \cite{casanueva2017benchmarking} for details. We extend this benchmark in three ways. Firstly, we add a new domain of recommending financial products. Secondly, we extend the user simulator to include context. Finally, we add our proposed algorithms and additional non-POMDP-based algorithms to the benchmark.

\subsection{Recommending Financial Products}
\label{subsec:fin_prods}
The financial domain is an interesting addition as it is different from domains currently in the benchmark: the number of interactions with a single user is typically limited, there may be large gaps in between interactions and user intentions are typically not constant over interactions. It is, for example, unlikely that a single customer needs multiple recommendations based on an intention to finance a car purchase. This renders approaches that require multiple interactions with a single user or that rely on direct estimation of user preferences inapplicable.

A second particularity of this domain is that different users have different familiarity with products. As a result, users in this domain have differing preferences and ability to express them. For example, customers that have a car loan will be more familiar with technicalities of secured loans and therefore be more capable of expressing their preferences for similar loans in detail. Such differences are common in domains with complex products, such as the financial, technology and automotive domains. Although the exact formulation of context is not the focus of this work and may vary per domain, we consulted with domain experts in the financial domain on contextual factors currently used in determining how to communicate with users across various channels. These domain experts indicate that one of the major factors in communicating about a product is whether the user consumes a product from the same product category.

\begingroup
\setlength{\tabcolsep}{2pt} 
\begin{table}[bthp]
\small
    \begin{tabular}{l|c|cc|cccc}
                                &               & \multicolumn{2}{c |}{\bf{Entropy}}     &   \multicolumn{4}{c}{\bf{POMDP}}          \\
                                & \rotatebox{90}{RQ}   &   \rotatebox{90}{EMDB}   &        \rotatebox{90}{EMDM}          &  \rotatebox{90}{HDC}         & \rotatebox{90}{RL$_{v}$}    & \rotatebox{90}{RL$_{s}$} & \rotatebox{90}{RL$_{bs}$}     \\ \hline
    Task-specific             &	  	        &	v  &       		            & 	                & 	                &               &                   \\
    NLU/DST-error aware   &		        &	 		&  	            		    & v            & v            & v        & v            \\          
    Adaptive                    &		        &	 		&  	        v		    &	                & v            & v        & v            \\      
    Uses context            &		        &	    	&  	    		            &	                &                   & fixed         &  adaptive         \\
    \end{tabular}
    \caption{Overview of qualities of approaches. RL$_v$, RL$_s$ and RL$_{bs}$ describe the vanilla, segmentation-based and belief-state based versions of $GP$, $A2C$, $DQN$ and $eNAC$.}
    \label{tab:benchmark}
\end{table}

Differences with other domains are not limited to typical interaction patterns, however: the item set $X$ is distinctive in this novel setting as well. This item set was developed using using well-known ontology engineering practices and evaluated with domain experts \cite{noy2001ontology} \cite{antoniou2004semantic}.
The resulting item set consists of 14 products and 13 features. Nine out of these can be used as a constraint by the user, see Table~\ref{tab:error_slots} for an overview. All other slots are only used to inform the user about the product and not relevant to the recommendation task. The number of values for all constraint features is 64. 
When compared to the existing domains in literature, the novel FIN domain has a relatively small item set and relatively large number of constraint-slots. We add this item set as an `ontology' to the Pydial benchmark for DM systems \cite{casanueva2017benchmarking} which is described in the next Sections in more detail.

\subsection{User Simulator}
\label{subsec:user_models}
We adapt the user simulator in the benchmark as described in \cite{schatzmann2007agenda} to reflect the scenario from the previous section. A full description of this simulator is out of scope and we limit ourselves to the main concepts before moving on to the extensions. In the simulator, actions by the simulated user are conditioned on the dialogue so far and on behavior parameters and includes an error model for ASR and NLU modules. Parameters for all of these have been tuned using data from experiments with real users, for details see \cite{schatzmann2007agenda}. Behavior parameters are sampled at the start of each dialogue and according to distributions that have been set in user profiles so that each dialogue is with a user with individual behavior characteristics. Similarly, up to three constraints $c_j$ are sampled randomly for each new simulated user. Additionally, heuristics to constrain the action space can be enabled or disabled. These \emph{action masks} make part of the action space unavailable and ease the learning task. A combination of user model, error model and availability of action mask is denoted as an `environment'. In total, the benchmark we use includes six different environments \cite{casanueva2017benchmarking}.

We extend the tuned simulator with user context to reflect the scenario from the previous section. Two user groups are modelled. The first group represents `laypersons' that express constraints for specific slots only; the second group represents knowledgeable users that express constraints for all slots.  All slots and their usage per group are listed in Table~\ref{tab:error_slots}. The usage of slots between groups for the FIN domain has been set after consultation with domain experts. For the CR, SFR and LAP domains, these are set to allow for a comparison of approaches across settings.

We add a $b_c$ to describe the user context and add per-slot constraint usage parameters to the simulator. Specifically, $b_c$ is a vector of two values, describing the belief on the user having experience in the domain or not. Although our approach facilitates a wide range of values, we here limit ourselves to the case of fully certain upfront knowledge, i.e. $b_c \in \{0,1\}^2$. We assume that interactions with both types of users are equally likely.

\subsection{Algorithms}
\label{sec:benchmark}
We evaluate our approach using all algorithms in the benchmark presented in \cite{casanueva2017benchmarking} and measure per-dialogue rewards according to equation~\ref{eq:reward} in Section~\ref{sec:rl4dialogue} across 10 random seeds with 4000 training and 500 test dialogues each. The benchmark contains one handcrafted policy, $HDC$, and four RL-based algorithms: $GP$ for GP-SARSA, $A2C$, $eNAC$ and $DQN$. All of these algorithms are based on the POMDP formalism introduced in Section~\ref{sec:rl4dialogue}. $GP$ is a data-efficient nonparametric value-based approach that uses Gaussian Processes to estimate $Q^{\pi}(b,a)$ from equation~\ref{eq:q_function} \cite{gavsic2010gaussian}. $DQN$ similarly estimates these $Q$ values using a neural network, i.e. it is a parametric approach \cite{su2017sample}
\cite{van2016deep}. $A2C$ and $eNAC$ are parametric algorithms that estimate the policy $\pi(b)$ as defined in equation~\ref{eq:q_function} directly, where $A2C$ estimates $Q(b,a)$ additionally \cite{fatemi2016policy}.
We refer to \cite{casanueva2017benchmarking} for more detail on these algorithms. We include vanilla versions of the learning algorithms, versions based on segmentation and versions based on an altered belief-state and denote these by $_v$, $_s$ and $_{bs}$ subscripts respectively.

We further extend the benchmark with three non-RL-based algorithms.\footnote{Code: \url{https://bitbucket.org/florisdenhengst/pydial/commits/tag/web-intelligence-19}} The algorithms were selected based on the task formalization of Section~\ref{sec:task} and to enable a comparison of learning algorithms versus handcrafted algorithms. Specifically, we add a randomized baseline, an algorithm with a search heuristic and a state-of-art learning method from \cite{wu2015entropy}. This last method keeps a history of successful dialogues as trajectories of user utterances $u$ and system actions $\langle u^1, a^1, ..., u^{\ell}, a^{\ell}\rangle$ up to a successful recommendation $a^{\ell}$. During a dialogue $\langle u^1, a^1, \ldots, u^t\rangle$, the system selects the action $a^{t}$ that minimizes the entropy of all past successful recommendations $a^{\ell}$, breaking ties with a random selection. We denote this approach with $EMDM$ for `Entropy Minimization Dialog Management'.

The two remaining non-POMDP-based algorithms are a randomized baseline and a baseline that uses information about the product database. The randomized baseline randomly asks for constraints on feature $f_j$ until there are no differentiating features in $X_{C_t}$ and then recommends some item $x_i \in X_{C_t}$ randomly. We denote this baseline with $RQ$ for `Random Question`. The second baseline has the same strategy for recommending an item, but differs in selecting $f_j$. Given the current $X_{C_t}$, it selects the $f_j$ with the highest entropy in the candidate item set $X_{C_t}$ and requests the user preference for it. This is a task-specific approach that uses a entropy as a heuristic to search the item set $X_{C_t}$ efficiently. We denote this benchmark as $EMDB$ for `Entropy Minimization DataBase`. All non-POMDP-based approaches, i.e. $RQ$, $EMDM$ and $EMDB$, have no way of dealing with errors from the ASR and NLU modules in Figure~\ref{fig:approaches}. The output of these modules with the highest confidence score is simply assumed as correct and used as input to these algorithms.

\subsection{Environment and Hyperparameters}
\label{sec:environment}
All experiments were run on Intel Xeon Silver 4110 Processors using Python version 2.7.9, TensorFlow version 1.12.0, NumPy version 1.15.4 and SciPy version 1.2.0. Ten different random seeds ranging from zero to ten were used. Hyperparameters were set as in \cite{casanueva2017benchmarking}, we repeat them here. For the $GP$ algorithm, a linear kernel was used on the state space and a Kronecker delta kernel was used on the action space. The `scale' variable of these determines the rate of exploration and was set to 3.

$DQN$, $A2C$ and $eNAC$ use an $\epsilon$-greedy exploration strategy during training where $\epsilon$ is linearly scaled between $\epsilon_s$ and $0.05$ in training, i.e. for the 4,000 dialogues. Exploration was turned off during evaluation. See Table~\ref{tab:hyperparams} for values of $\epsilon_s$ and network architecture for the neural network based approaches. For these, the architecture consisted of three layers of fully connected feedforward of varying sizes. The Adam optimizer was used for training with an initial learning rate of $0.001$. We refer to the code repository for further details on the hyperparameters.

\begin{table}[bthp]
    \begin{tabular}{l c c c c}
          & \multicolumn{2}{c}{\# Nodes} & \\
    Model & Hidden Layer 1  & Hidden layer 2    & $\epsilon_s$ \\ \hline
    $DQN$ & 300             &   100             & .5 \\
    $A2C$ & 200             & 75                & .5 \\
    $eNAC$ &130             & 50                & .3 \\ \hline
    \end{tabular}
    \caption{Hyperparameters for neural network based approaches.}
    \label{tab:hyperparams}
\end{table}

\section{Results}
\label{sec:results}
In this section, we describe the results with respect to the research questions from Section~\ref{sec:setup}.
Table~\ref{tab:rewards} lists all results.


\textbf{Q1} Figure~\ref{fig:results}a shows the performance of the best algorithms in an environment where ASR/NLU errors are absent. According to hypothesis H1, we expected the $HDC$ and $EMDB$ algorithms to outperform learning algorithms. We analyse the performance of these algorithms per domain. The CR domain contains relatively little slots and groups are similar. The task-specific $EMDB$ algorithm moderately outperforms learning-based approaches $GP_s$ and $DQN_s$ which in turn outperform the $HDC$ algorithm. Moving to the FIN domain, $DQN_s$ and $GP_s$ outperform $HDC$ due to the large difference between groups. We analyze the poor results of $EMDB$ in this novel domain below (Q3). In the LAP domain, the $EMDB$ algorithm performs the worst out of the selected algorithms. This domain has a large number of slots hence there is likely to be a differentiating feature $f_j$ that will be selected according to $EMDB$. The $EMDB$ algorithm thus keeps on asking for new $f_j$, even when the user has already listed all of their requirements. Comparing $HDC$ with learning-based approaches in this domain, it performs comparable to $DQN_s$ and $GP_s$. The reason for this may be that this is a relatively challenging learning task which limits the benefits of personalization. The SFR domain has a relatively large item set $X$ and a moderate number of slots. The search heuristic of $EMDB$ works as expected here and $GP_s$ and $DQN_s$ moderately outperform handcrafted approaches. Overall, we find that --in contrast to H1-- learning-based approaches perform comparable or better than both handcrafted approaches, \emph{even in the absence of ASR/NLU errors}.

We now compare these families of approaches in an environment with ASR/NLU errors in Figure~\ref{fig:results}b. In this setting, the gold standard $HDC$ algorithm degrades more than learning approaches, further supporting the benefits of learning approaches in a scenario with different user groups. The difference can be explained by $HDC$'s response to an unclear answer for some slot: it requests the user to confirm the most likely value as recorded by the ASR/NLU modules. Such a request will not further the dialogue if that particular slot does not contain a constraint for the user. The $HDC$ algorithm does not take this into account, whereas learning approaches can adapt to the laypersons' inability to informatively respond after such a confirmation request and ask for other constraints first. The $EMDB$ algorithm cannot handle uncertainty from ASR/NLU outputs. It assumes the most likely preference as indicated by ASR/NLU modules. This assumption is occassionally incorrect and generally ruins $EMDB$'s performance.

\begin{figure*}[htbp]
    \centering
    \includegraphics[width=.8\textwidth]{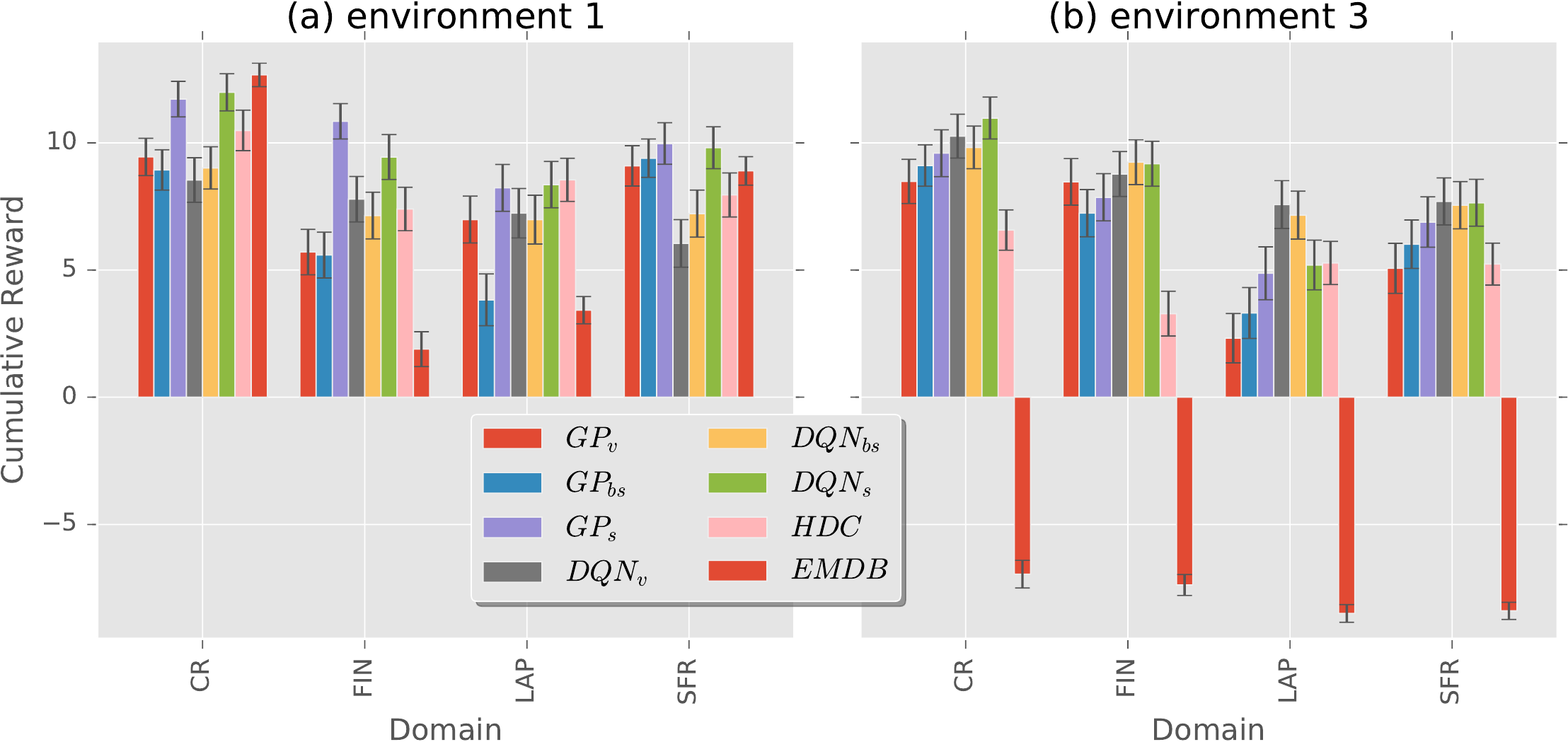}
    \caption{Average reward per dialogue in test set for environments without (a) and with (b) ASR/NLU errors.}
\label{fig:results}
\end{figure*}

\textbf{Q2} In contrast to hypothesis H2, performance of belief state- and segmentation-based personalization approaches vary across domains, environments and used learning algorithms. For the $GP$ algorithm, segmentation generally outperforms vanilla and belief-state based approaches in both environments. This suggests that $GP$ suffers less from lack of training data as a result of segmentation, which is in line with earlier findings that $GP$ is a data efficient algorithm \cite{gavsic2010gaussian}. The performance of this algorithm relies on the chosen kernel. In the benchmark, a linear kernel is used. This kernel assumes a linear relation between $Q^{\pi}(b,a)$ and the belief state $b$. We briefly analyze this linearity assumption by considering two similar belief states $b$ that only differ in the belief on user group membership for the current user $b^c$. The linearity assumption implies that some favorable action for the first group is unfavorable for the other group. This assumption clearly does not hold for some actions, e.g. requesting some $f_j$ that is used by both groups.

For $DQN$, some negative effects of segmentation can be seen in cases with a complex learning problem, i.e. in environments with ASR/NLU errors and in domains with a large state space. These negative effects can be mainly seen in domains with larger state spaces LAP and SFR. Regarding the belief state-based approach, results indicate that it performs comparable or slightly better than the vanilla approaches in most configurations. We hypothesized that this approach would learn to exploit differences in user population without suffering from the drawback of limited training data as in the segmentation-based approach. Although our findings indicate that the latter is generally the case, the benefits of personalization diminish for more complex learning problems in environments 4-6. A possible explanation for this is that the algorithms' hyperparameters, specifically the neural network architecture for $DQN$ and kernel for $GP$, were not optimized to the personalization setting.

\textbf{Q3} Figure~\ref{fig:domain} shows how POMDP-based approaches hold over various domains in all included environments. We omit non-POMDP-based approaches here due to their poor performance in environments 3-6. When comparing the novel FIN domain, the gold standard $HDC$ is outperformed by all considered learning algorithms. The learning algorithms generalize to the new domain. The $HDC$ policy was handcrafted for the other four domains and does not transfer well to a novel domain with different characteristics. To analyze the results of $EMDB$ in the FIN domain, we consider again Figure~\ref{fig:results}. In the FIN domain, the item set $X$ is small which makes the search heuristic on which $EMDB$ relies inapplicable. These results are in line with hypotheses H3a and H3b.

\begin{figure}[htbp]
    \centering
    \includegraphics[width=1\columnwidth]{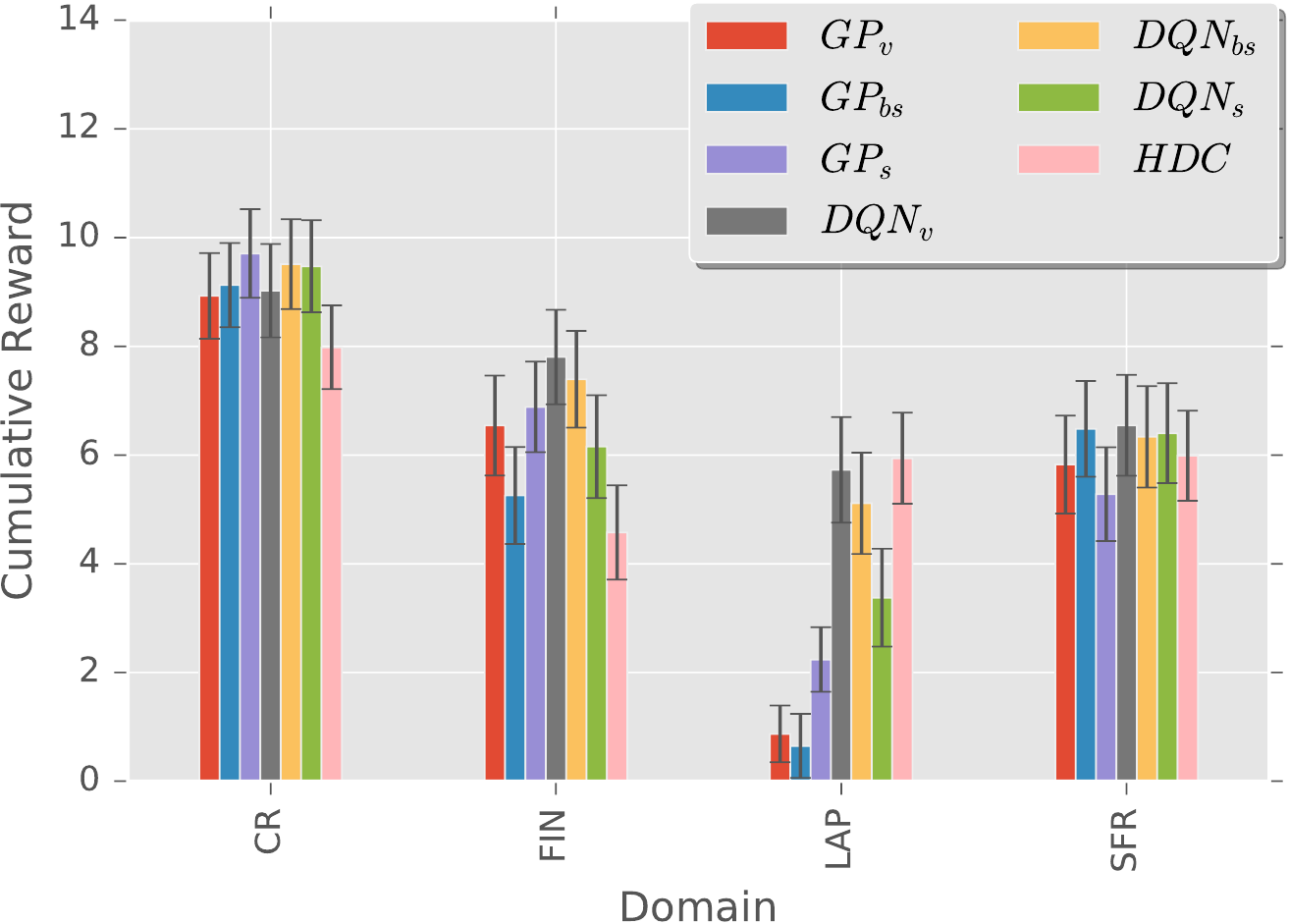}
    \caption{Per-dialogue reward of selected algorithms in test set, averaged over all environments.}
    \label{fig:domain}
\end{figure}

\begingroup
\setlength{\tabcolsep}{3pt} 
\begin{table*}[htbp]
\small
    \begin{tabular}{l c c c c | c c c | c c c | c c c | c c  c | c | c | c | c }
    
       \rotatebox{90}{env} &
       \rotatebox{90}{Error Model} &
       \rotatebox{90}{Action Masks} &
       \rotatebox{90}{User Model} &
       \rotatebox{90}{domain} & \rotatebox{90}{$A2C_v$} & \rotatebox{90}{$A2C_{bs}$} & \rotatebox{90}{$A2C_s$} & \rotatebox{90}{$DQN_v$} & \rotatebox{90}{$DQN_{bs}$} & \rotatebox{90}{$DQN_s$} & \rotatebox{90}{$eNAC_v$} & \rotatebox{90}{$eNAC_{bs}$} & \rotatebox{90}{$eNAC_s$} & \rotatebox{90}{$GP_v$} & \rotatebox{90}{$GP_{bs}$} & \rotatebox{90}{$GP_s$} & \rotatebox{90}{$HDC$} & \rotatebox{90}{$RQ$} & \rotatebox{90}{$EMDB$} & \rotatebox{90}{$EMDM$} \\ \hline
    \multirow{4}{*}{1} &
    \multirow{4}{*}{0\%} &
    \multirow{4}{*}{on} &
    \multirow{4}{*}{
        \rotatebox{90}{normal}
    } &
     CR  &   12.2    &   11.6    &   10.6    &   8.5     &   9.0 &   12.0&   1.4     &   6.5 &   10.6    &   9.4 &   8.9 &   11.7    &   10.5    &   \textbf{12.7}&   \textbf{12.7}    &   -4.7\\ 
    & & & &  FIN &   \textbf{10.9}    &   8.5     &   7.2     &   7.8     &   7.1 &   9.4 &   4.1     &   0.8 &   8.0     &   5.7 &   5.6 &   10.8    &   7.4     &   2.3 &   1.9     &   -12.3\\ 
    & & & &  LAP &   5.9     &   4.1     &   0.6     &   7.2     &   7.0 &   8.4 &   7.5     &   7.9 &   5.9     &   7.0 &   3.8 &   8.2     &   \textbf{8.5}     &   4.2 &   3.4     &   -14.0\\ 
    & & & &  SFR &   6.4     &   6.4     &   5.1     &   6.0     &   7.2 &   9.8 &   5.2     &   5.0 &   8.3     &   9.1 &   9.4 &   \textbf{10.0}    &   8.0     &   9.2 &   8.9     &   -8.8\\ 

    \hline
    \multirow{4}{*}{2} &
    \multirow{4}{*}{0\%} &
    \multirow{4}{*}{off} &
        \multirow{4}{*}{
        \rotatebox{90}{normal}
    } &
     CR  &   2.8     &   2.3     &   2.2     &   11.8    &   11.2&   11.3&   -4.4    &   -3.8&   3.2     &   11.7&   11.6&   11.7    &   11.9    &   \textbf{12.7}&   \textbf{12.7}    &   -4.7\\ 
    & & & &  FIN &   2.8     &   3.2     &   3.4     &   10.7    &   9.8 &   5.7 &   -3.2    &   -2.2&   3.8     &   8.1 &   5.4 &   6.7     &   8.5     &   2.3 &   -10.0   &   -12.3\\ 
    & & & &  LAP &   -2.7    &   -2.4    &   -2.5    &   6.3     &   5.7 &   1.8 &   -3.3    &   -3.7&   -0.1    &   -1.0&   -0.9&   -0.9    &   10.3    &   4.2 &   3.4     &   -14.0\\ 
    & & & &  SFR &   -0.8    &   0.1     &   -1.6    &   9.4     &   7.4 &   7.4 &   5.0     &   5.1 &   0.2     &   8.8 &   8.6 &   5.4     &   10.3    &   9.2 &   8.9     &   -8.8\\ 

    \hline
    \multirow{4}{*}{3} &
    \multirow{4}{*}{15\%} &
    \multirow{4}{*}{on} &
        \multirow{4}{*}{
        \rotatebox{90}{normal}
    } &
    CR  &   8.2     &   8.1     &   7.8     &   10.3    &   9.8 &   \textbf{11.0}&   7.0     &   8.0 &   10.0    &   8.5 &   9.1 &   9.6     &   6.6     &   -7.4&   -7.0    &   -5.3\\ 
    & & & &  FIN &   6.2     &   5.4     &   3.2     &   8.8     &   \textbf{9.2} &   \textbf{9.2} &   4.6     &   7.0 &   6.8     &   8.5 &   7.2 &   7.9     &   3.3     &   -7.8&   -7.4    &   -12.5\\ 
    & & & &  LAP &   -1.3    &   -0.9    &   -2.2    &   \textbf{7.6}     &   7.2 &   5.2 &   5.7     &   5.5 &   4.6     &   2.3 &   3.3 &   4.9     &   5.3     &   -8.7&   -8.5    &   -14.3\\ 
    & & & &  SFR &   0.8     &   1.1     &   0.1     &   \textbf{7.7}     &   7.5 &   7.6 &   6.3     &   7.1 &   4.2     &   5.1 &   6.0 &   6.9     &   5.2     &   -8.4&   -8.4    &   -9.7\\ 

    \hline
    \multirow{4}{*}{4} &
    \multirow{4}{*}{15\%} &
    \multirow{4}{*}{off} &
        \multirow{4}{*}{
        \rotatebox{90}{normal}
    } &
     CR  &   2.4     &   2.6     &   1.4     &   \textbf{10.2}    &   9.5 &   7.1 &   0.9     &   1.7 &   2.9     &   9.6 &   9.7 &   8.9     &   6.6     &   -7.4&   -7.0    &   -5.3\\ 
    & & & &  FIN &   3.3     &   4.2     &   1.3     &   \textbf{9.6}     &   7.1 &   4.6 &   -1.0    &   -1.0&   4.3     &   6.4 &   5.2 &   5.4     &   3.3     &   -7.8&   -7.4    &   -12.5\\ 
    & & & &  LAP &   -3.0    &   -3.1    &   -2.7    &   4.6     &   3.4 &   -0.1&   -3.8    &   -0.3&   -2.5    &   -1.1&   -1.0&   -1.0    &   \textbf{5.3}     &   -8.7&   -8.5    &   -14.3\\ 
    & & & &  SFR &   -1.0    &   0.2     &   -1.8    &   5.2     &   \textbf{6.7} &   4.3 &   -1.1    &   2.0 &   0.9     &   4.6 &   4.7 &   2.5     &   5.2     &   -8.4&   -8.4    &   -9.7\\ 
    \hline
    \multirow{4}{*}{5} &
    \multirow{4}{*}{15\%} &
    \multirow{4}{*}{off} &
        \multirow{4}{*}{
        \rotatebox{90}{unfriendly}
    } &
     CR  &   6.6     &   4.6     &   4.8     &   7.0     &   \textbf{9.7} &   8.3 &   4.9     &   7.7 &   7.6     &   7.5 &   8.3 &   8.9     &   6.7     &   -7.5&   -7.5    &   -5.5\\ 
    & & & &  FIN &   2.2     &   2.1     &   1.6     &   6.2     &   \textbf{7.2} &   4.1 &   4.4     &   5.5 &   5.1     &   5.3 &   4.6 &   5.6     &   2.5     &   -7.8&   -7.5    &   -12.8\\ 
    & & & &  LAP &   -3.3    &   -2.0    &   -3.1    &   3.7     &   \textbf{4.1} &   1.9 &   1.8     &   1.8 &   0.5     &   -0.0&   -0.1&   1.7     &   3.0     &   -8.6&   -8.4    &   -14.6\\ 
    & & & &  SFR &   -2.1    &   -0.1    &   -1.1    &   \textbf{5.3}     &   4.6 &   4.6 &   2.3     &   3.3 &   4.1     &   3.8 &   3.6 &   3.5     &   3.7     &   -8.4&   -8.4    &   -10.3\\ 
    \hline
    \multirow{4}{*}{6} &
    \multirow{4}{*}{30\%} &
    \multirow{4}{*}{on} &
        \multirow{4}{*}{
        \rotatebox{90}{normal}
    } &
    CR  &   4.2     &   4.2     &   4.8     &   6.4     &   \textbf{7.8} &   7.2 &   6.2     &   7.1 &   7.2     &   6.8 &   7.1 &   7.3     &   5.6     &   -4.7&   -4.7    &   -5.8\\ 
    & & & &  FIN &   0.6     &   0.2     &   0.5     &   3.7     &   3.8 &   3.8 &   3.8     &   4.8 &   \textbf{5.6}     &   5.2 &   3.5 &   4.9     &   2.5     &   -7.6&   -7.0    &   -12.6\\ 
    & & & &  LAP &   -2.8    &   -2.6    &   -2.3    &   \textbf{4.9}     &   3.4 &   3.1 &   3.2     &   3.3 &   2.0     &   -2.0&   -1.2&   0.4     &   3.2     &   -9.3&   -8.8    &   -14.5\\ 
    & & & &  SFR &   1.6     &   -1.8    &   -0.5    &   \textbf{5.7}     &   4.6 &   4.6 &   4.2     &   4.7 &   4.9     &   3.6 &   2.4 &   3.5     &   3.5     &   -8.3&   -8.0    &   -9.7\\ 
     \hline
    \multicolumn{5}{c|}{mean}
    & 2.51
	& 2.34
	& 1.54
	& 7.28
	& 7.09
	& 6.35
	& 2.57
	& 3.49
	& 4.52
	& 5.54
	& 5.38
	& 6.03
	& 6.12 
	& -2.92
	& -3.37
	& -10.38
    \end{tabular}
    \caption{Average reward per dialogue for test set across environments, domains and algorithms in the benchmark.}
    \label{tab:rewards}
\end{table*}

\section{Discussion}
\label{sec:discussion}
In this work, we have proposed two approaches to DM using personal context and evaluated them on various environments, in various domains and using various algorithms. The approaches leverage existing contextual information about a particular user and can offer personalized DM even in the absence of previous interactions with a particular user.

In order to evaluate our approaches, we have extended an existing benchmark for conversational item recommendation with two user contexts and associated behavior patterns. The behavior patterns reflect those found in domains where `expert' and `layperson' users have differing knowledge about the available items. Results indicate that learning a dialogue policy is beneficial in settings with differing user behaviors. Notably, the addition of context boosts performance of learned dialogue managers to comparable or higher levels than a handcrafted gold standard and task-specific approaches, even in an environment without noise from preprocessing modules.

We find that performance of learning approaches varies with environment, domain, and algorithm. Specifically, data efficiency could be investigated by increasing the number of training dialogues. Similarly, the applicability of the approaches could be investigated by varying the difference between user groups. Furthermore, varying hyperparameter settings such as neural network architecture and learning rate and more powerful and stable RL algorithms may lead to more the complex behaviors in the new setting such as those in \cite{haarnoja2018soft}. More experiments are necessary to further investigate performance characteristics for the proposed approaches.

With regards to methodology, we have introduced a case validated by domain experts in the financial domain and added it to an existing benchmark of item recommendation. We have extended a realistic user simulator with additional behavior parameters for all domains in the benchmark to comprehensively test our approaches. Although these additional parameters are suitable to test our approaches technically, they were not sampled from real-world data. Comparing the approaches in real-world settings, such as an evaluation with real users or an evaluation in a configuration where behavior parameters are based on real-world differences between experts and laypersons would be interesting next steps.

Finally, we tested our approaches to the usage of context in a specific case with different user groups with static context information and a constant action space. Our approaches, however, are general and could be applied to various other usages of context to dialogue policy optimization. Especially interesting would be the inclusion of sentiment estimates as in \cite{poria2016fusing}. Together with an extension of the action space, these could aid in making the conversation more natural by conditioning e.g. trust-building system responses on conversation content and context at the same time.

\bibliographystyle{plain}
\bibliography{main}

\end{document}